\documentclass[letterpaper, 10 pt, conference]{ieeeconf}
\IEEEoverridecommandlockouts              
\usepackage{graphics}
\usepackage{epsfig} 
\usepackage{mathptmx}
\usepackage{times} 
\usepackage{amsmath}
\usepackage{amssymb}
\usepackage{cite}
\usepackage{here}
\usepackage{xcolor}
\usepackage{bm}
\usepackage{algorithm}
\usepackage{algorithmic}
\usepackage{array}

\title{\LARGE \bf
 Mapless Navigation: Learning UAVs Motion for\\ Exploration of Unknown Environments}
\author{Sunggoo Jung$^{1}$ and David Hyunchul Shim$^{2}$
\thanks{This paper is based upon work supported by the Ministry of Trade, Industry \& Energy (MOTIE, Korea) under the Industrial Technology Innovation Program. No.10067202, "Development of Disaster Response Robot system for lifesaving and Supporting Fire Fighters at Complex Disaster Environment"}%
\thanks{$^{1}$Sunggoo Jung is with the Autonomous UAV Research Section, ETRI, Daejeon, Korea. {\tt\small sunggoo@etri.re.kr}}
\thanks{$^{2}$David Hyunchul Shim is with the Unmanned Systems Research Group (USRG), KAIST, Daejeon, Korea. {\tt\small hcshim@kaist.ac.kr}}%
}

\begin{document}
\maketitle
\thispagestyle{empty}
\pagestyle{empty}
\begin{abstract}
This study presents a new methodology for learning-based motion planning for autonomous exploration using aerial robots. Through the reinforcement learning method of learning through trial and error, the action policy is derived that can guide autonomous exploration of underground and tunnel environments. A new Markov decision process state is designed to learn the robot's action policy by using simulation only, and the results is applied to the real-world environment without further learning. Reduce the need for precision map in grid-based path planner and achieve map-less navigation. The proposed method can have a path with less computing cost than the grid-based planner but has similar performance. The trained action policy is broadly evaluated in both simulation and field trials related to autonomous exploration of underground mines or indoor spaces.
\end{abstract}

\section{INTRODUCTION}
As powerful computations become possible on small computers, there is a growing demand to combine them with drones to explore areas that are difficult for humans to reach\cite{bircher2016three, shakhatreh2019unmanned}. Following this trend, the DARPA Subterranean Challenge requires robotic agents to explore various underground spaces ridden with obstacles to find a set of objects including human victims for rescue. Such a mission will require drones to find its location and a collision-free path in real-time using local maps built \textit{on the fly}. Autonomous exploration in these environments poses a series of challenging problems at many levels. GPS signals may not be available\cite{papachristos2019autonomous}, and may be a condition in which only limited lighting exists\cite{kim2019autonomous}. In addition, flight between narrow gaps and low heights of the ceiling may be required\cite{falanga2017aggressive}.\\
In general, method for autonomous exploration in unknown and cluttered environments using aerial robots is as follows: 1) generate a map of the surrounding environment from gathered sensor information; 2) convert the generated map into a path plan-appropriate format such as a Probability grid; 3) applying path planning techniques such as A* or RRT*; 4) extract waypoint from generated path; 5) generate control command to track the waypoint information. However, problem of the traditional autonomous exploration approach is that the key exploration strategy depends heavily on precise maps, so the computational burden increases rapidly as the area of exploration increases\cite{julia2012comparison}.\\
To address these issues, we propose a mapless autonomous exploration based on deep reinforcement learning (DRL), a novel motion planning algorithm that exploits LiDAR measurements to recognize the spatial data of the surrounded environment around the aerial robot. Define the status of the Markov decision process (MDP) using the ambient data of the LiDAR point cloud, the current pose of vehicle and the target location to be reached. The proposed framework can bridge the reality gap between synthetic simulations and physical robots. Without further learning, the results of the policies learned in the simulation can be applied to the real-world environment and it shows performance similar to traditional grid-based planning with less computing power.
\begin{figure}[!t]
    \centering
    \includegraphics[width=3.2in]{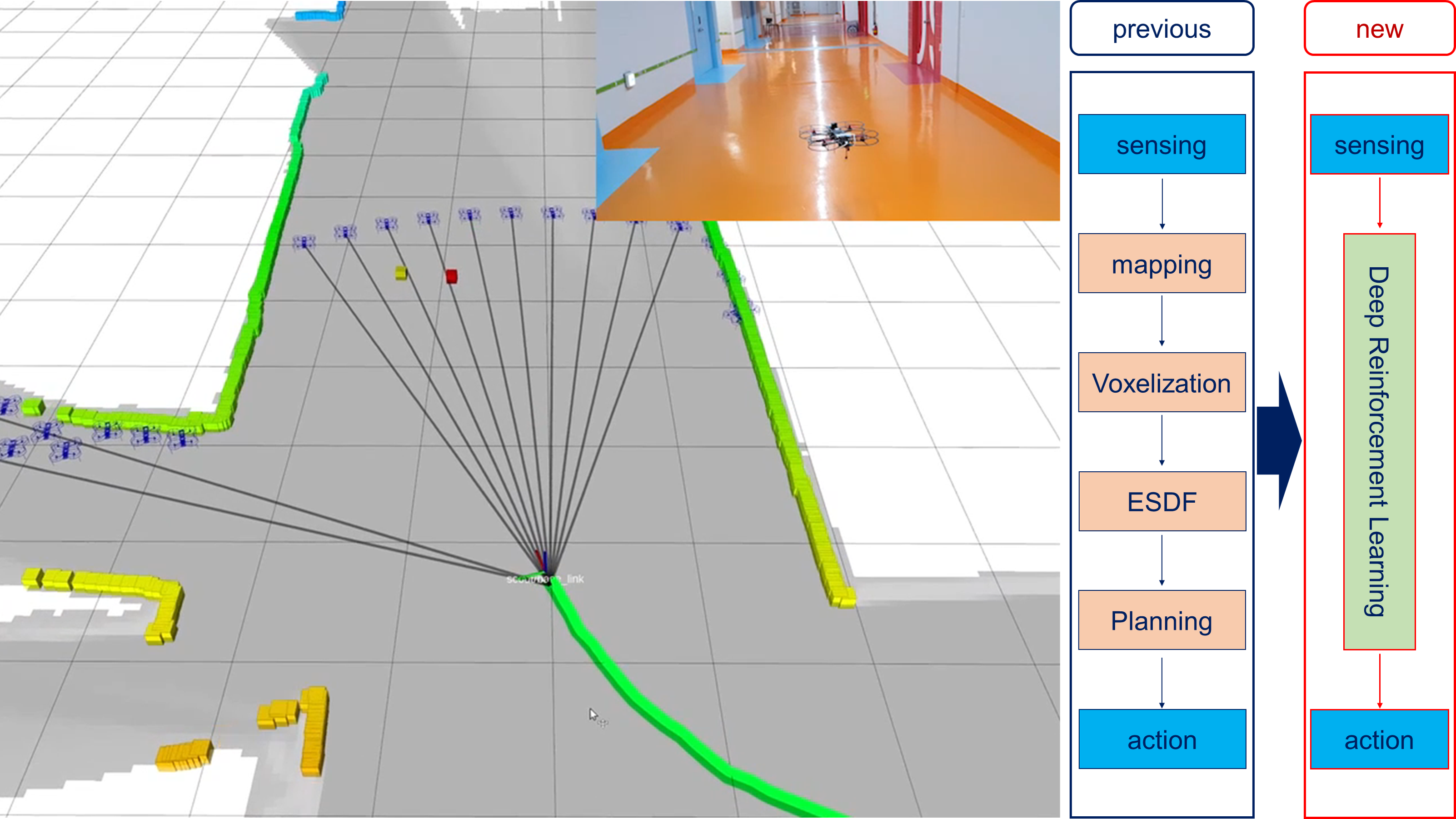}
    \caption{A snapshot of autonomous flight using propose DRL based exploration framework in a real-world environment}
    \label{fig:MAV}
\end{figure}
The main contributions of this paper are:
\begin{itemize}
\item We propose a DRL based autonomous exploration approach by utilizing a new MDP state which can allow us to use the simulation learning results directly in an real-world experiment without further learning
\item We newly use collision-free open space end-point selection of a local segment path to take an obstacle-aware motion planning
\item Demonstration of the proposed method in a fully onboard quadrotor system. We evaluate the DRL result on both simulation and real-world by traversing a large building and subterranean environment
\end{itemize}

\section{RELATED WORK}
\label{related_work}
Various approaches for autonomous exploration of an unknown area using UAVs have been proposed and implemented \cite{meera2019obstacle}, \cite{kuipers1991robot}, \cite{liu2018search}. Early work includes frontiers-based exploration \cite{yamauchi1997frontier}, model predictive control (MPC) based path re-planning experimental work \cite{shim2005autonomous} also presented. Herein, we provide a brief overview of state-of-the-art contributions that are especially relevant to exploring unknown/unexplored area using UAVs.
\subsection{Online Path Re-planning of UAVs}
Unknown area exploration planner was proposed by \cite{bircher2016receding}. This work successfully shows the capability of priory unknown area exploration while collision avoidance. Kino-dynamic real-time planning using quadrotor is presented in \cite{allen2016real}. This shows full-stack, planning architecture, however the state-estimation is relay on the VICON motion capture system. Front-end path planning and back-end path smoothing works have proposed in \cite{gao2017gradient} and show impressive results in a small range of indoor and outdoor area. MPC based online path planning work show dynamic re-planning with considering multi-object concept\cite{peng2012intelligent}, however the work finished within simulation without real-world validation. One line segment combination path smoothing approach is presented in \cite{lai2016robust}. This work generates the path using a two-point boundary value problem approach to reduce the computation cost. This work showed reliable flight result, however, the piece-wise segment of path joint optimization is not presented. Interesting works is presented by \cite{popovic2016online}. In this work, they developed an online path planning algorithm to reduce the energy cost while detecting the weeds. They showed impressively reducing the energy cost by effective path planning algorithm but still also remain in simulation.
\subsection{DRL based exploration}
Due to the recent significant development of DRL, researchers in the robot control area have tried many studies linking generated policy from DRL results to robot control. Modular network is used for Successful Robot navigation \cite{tai2016mobile, walker2019deep}. However, these methods simply focus on flying without collisions. For autonomous exploration, \cite{li2019deep} combine traditional navigation approaches with DRL-based decision algorithm and design additional segmentation network architecture. \cite{niroui2019deep} integrates DRL with frontier exploration. While these methods demonstrate successful exploration performance in cluttered area, gridding map information must always be exist. \cite{reinhart2020learning} propose imitation learning from traditional graph-based planner. Although it showed good performance, it requires further learning in the real world.
\section{PROBLEM STATEMENT}
\label{problem_statement}
The DRL-based end-to-end control method has the advantage of being concise, but it requires a large trial and error experience. Therefore, this learning is generally done in the simulation environment. However, due to the reality-gap, it is difficult to bring the results learned from the simulation into the real-world. Based on these issues, the following methods are used in this paper.
\begin{enumerate}
\item Improve learning convergence by replacing navigation with the problem of planning a short-distance path from your current location to a selected local goal. This local-goal is planned as a receding horizon planning strategy to expand the exploration area.
\item In order to use the learned policy under simulation directly into the real environment, we define the MPD state with superior generalization.
\item Map the learned motion to the high-level Velocity controller of the robot so that it is not related to the dynamics of the actual robot.

\end{enumerate}

\section{METHODOLOGY}
\label{method}
The MDP state is designed to use TD advantageous actor-critic (A2C) reinforcement learning framework for unknown area exploration\cite{silver2014deterministic, mnih2016asynchronous}. Initialize state $s_{0}$ according to the initial condition. Based on the initial state and policy, the algorithm selects the control action $a_{0}\in A$ ($A$ is \textit{action space} and there are eight actions in total which is called \textit{king's move}). As a result of the selection, the system randomly transitions to any new state $s_{k}$ and receives a reward according to the state transition probability $P_{s_{k-1}}^{a_{k-1}}(·)$. The main purpose of RL in this paper is to update policy through the MDP information to find the optimal collision-free path from current position to local-goal. 
\subsection{Local-goal selection}
\begin{figure}[!t]
    \centering
    \includegraphics[width=3.4in]{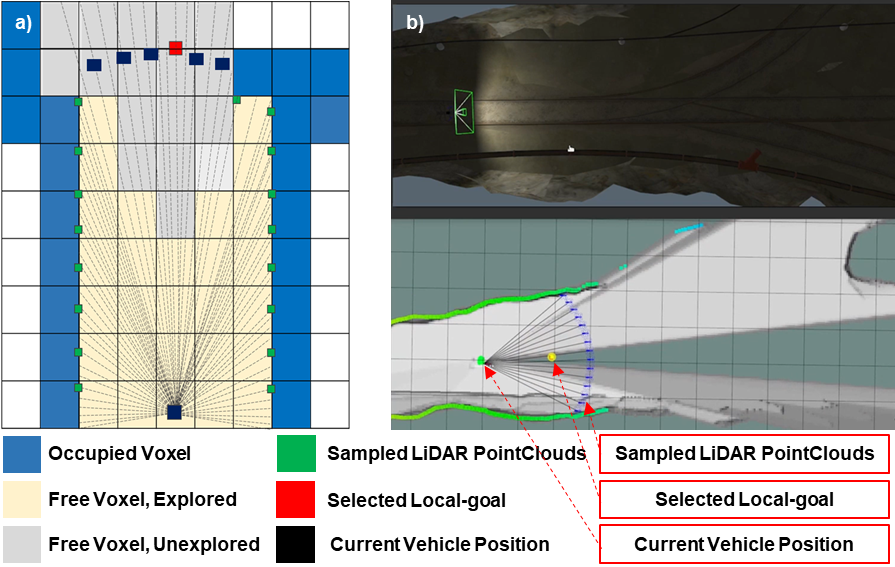}
    \caption{a) Conceptual figure of end-point selection from LiDAR range measurements. Evenly 48 point cloud is sampled around the vehicle. b) Example of a Gazebo simulation showing that a local goal is being selected in real time, depending on the current state of the vehicle.}
    \label{fig:endpoint}
\end{figure}
The LiDAR sensor measurements give the index and distance of the point which can be denoted by $[k, l_k]$. To select local-goal, we sample evenly 48 points out of 1440 LiDAR point clouds. The samples are at 30 intervals starting from the 1st index of the LiDAR among a total of 1440 LiDAR points(see, Fig. \ref{fig:endpoint}). This constitutes a set of 48 sample point clouds surrounding the vehicle at 270 degrees. Now, we collect a set of point clouds exceed a certain distance $d_ {th}$ (local-goal candidates, First line of Eq. \ref{eq:thres}). Find the average index of this set and convert the point cloud of the average index to the body-frame coordinate system. Finally we define the converted point as local-goal (Eq. \ref{eq:goal}.)
\begin{equation}
\label{eq:thres}
L_k = \left\{\begin{matrix} 4m -w_{\psi} & if\;\; \sigma_k = \infty,\;\; or \;\;\sigma_k >= d_{th}\:\;\;\;\;\;\;\;\;\;\;\;\;\;\;\\[2mm]
3m-w_{\psi} & if\;\; 3m \leq\sigma_k\leq d_{th},\\[2mm]
\sigma_k & otherwise,\;\;\;\;\;\;\;\;\;\;\;\;\;\;\;\;\;\;
\end{matrix}\right.
\end{equation} 
\begin{equation}
\label{eq:goal}
setb_{k}^{*}=\underset{k\in N}{argmax}\;\;\bm{x} + L_k\cdot \binom{cos(\psi+k\cdot\frac{\Omega}{N}-\frac{\Omega}{2})}{sin(\psi+k\cdot\frac{\Omega}{N}-\frac{\Omega}{2})}
\end{equation}
where $w_{\psi}$ is weight term to select index which is most perpendicular to the vehicle heading. Vehicle current position $\bm{x}$, heading $\psi$, and the sampled LiDAR point cloud pose $L=\{l_k\}_{k=1,...,K},l_k\in \mathbb{R}{^{2}}$ is in vehicle body-frame coordinate. $\Omega$ is the sample field of view, and $N$ is total number of samples. The selected $b_{k}^{*}$ is defined as the local-goal.

\subsection{MDP state design: narrowing the reality gap}
\label{narrow}
Methods for learning control commands has a disadvantage that the network becomes large because an encoder and a decoder must be used. Instead of learning the control command directly, we generalize the sensor data and replace it with the problem of learning the action in grid world.\\
\begin{figure}[!t]
    \centering
    \includegraphics[width=3.4in]{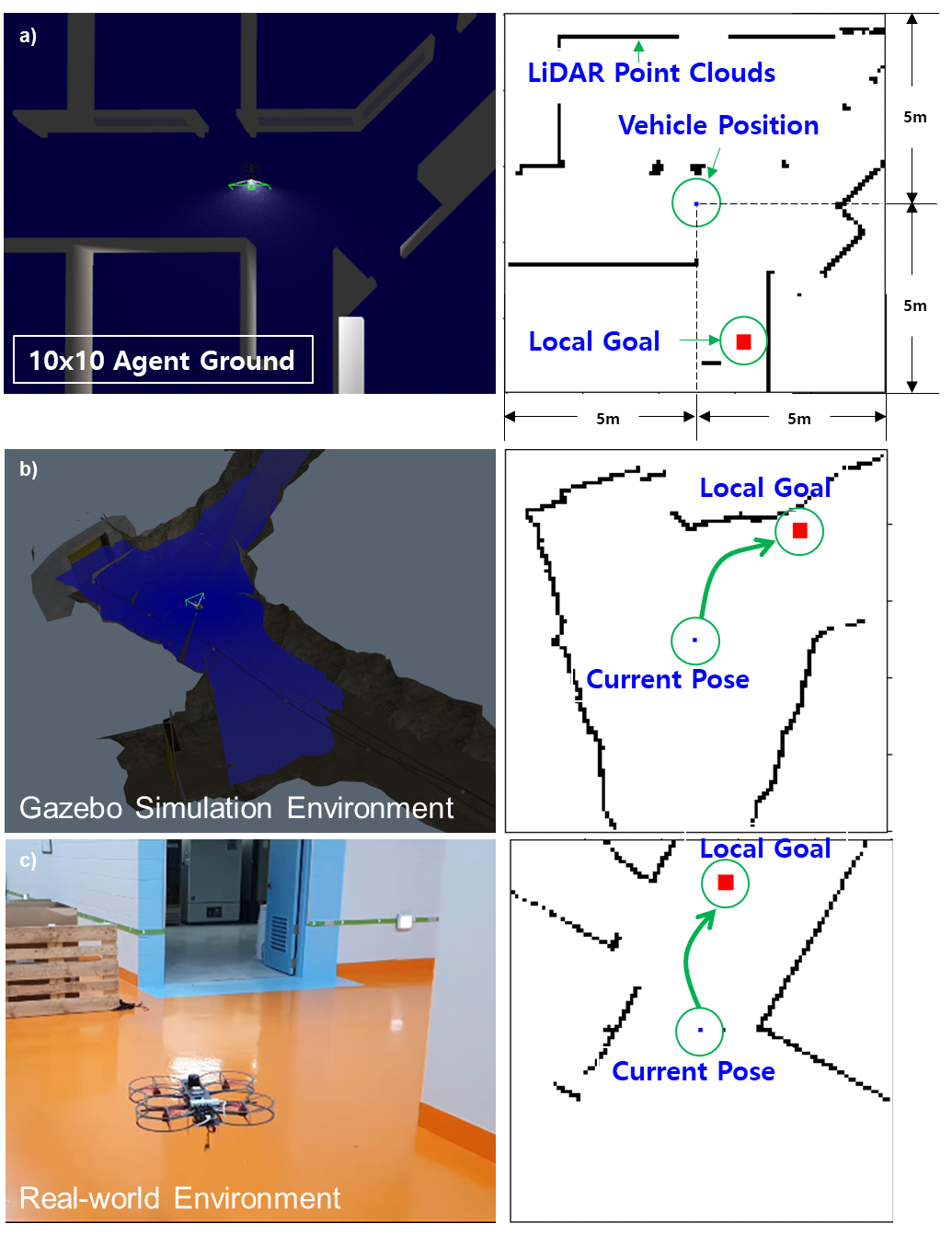}
    \caption{The agent ground and designed MDP states. State is configured LiDAR point cloud (black dots), vehicle current pose (blue dot), and local-goal (red dot)}
    \label{fig:agent_ground}
\end{figure}
We configure the space for the agent to learn. The local search space with a grid of 10cm in the space of 10m $\times$ 10m of the real-world is set as the agent ground (see Fig. \ref{fig:agent_ground} (a)). On this agent ground the LiDAR point cloud, the current position of an agent, and the target position is stored in the form of world coordinates. Therefore, the RL problem becomes a path-planning problem from current pose to local-goal in 100pixels $\times$ 100pixels grid world (see Fig. \ref{fig:agent_ground} (b),(c)). The left image shows the flight environment. Right image is the agent ground. It includes the LiDAR point in black color, the vehicle's current position in blue, and the target position, which is the local target that the agent to be reached, is marked in red. This eliminates the synthetic error of sensor information, reducing the gap with the real-world. The agent ground shown in this way is used as the input to CNN. Through this, feature points is extracted and used these result as an actor-critic input.

\subsection{Reward function: for collision-free exploration}
Creating the shortest path is important to efficiently explore unknown environments, but creating collision-free paths is more important. The reward function is designed as
\begin{equation}
r=\left\{\begin{matrix} 10\;\;\; (p(x_{1},y_{1})=g(x,y))
\\[2mm] |l_{prev}|-|l|\;\;\;\bigl(\begin{smallmatrix} (p(x_{1},y_{1})\neq g(x,y))
\\[2mm]  (p(x_{1},y_{1})\neq o(x,y))
\end{smallmatrix}\bigr) 
\\ -100\;\;\; (p(x_{1},y_{1})=o(x,y))
\end{matrix}\right.
\end{equation}  
where $p$ is the current position, $g$ is a local-goal, and $o$ represents ans obstacle which is colored in black in agent ground. If agent is not in both local goal and obstacle the reward is given by the temporal difference. Where $l$ is the euclidean distance from agent position to local-goal. Through this reward function, the agent gradually learns how to reach the target point with the best path while avoiding obstacles as learning progresses to maximize cumulative reward.

\section{The Neural Network in TD Actor-Critic}
The components of MDP is a tuple $(S, s_{0},A,{P_{s}^{a}},\gamma,R)$. In this paper, $S=\{s_{k}\}_{k=1,...,k,} s_{k}\in \mathbb{R}{^{2}}$ is the set of states including position of each lidar point cloud, current position of a drone and target point location. The network architecture builds along  the  same  lines  as TD Actor-Critic \cite{mnih2016asynchronous} which minimize the gradient:
\begin{equation} 
\label{eq:acTD} 
\nabla_{\theta}J(\theta)=\mathbb{E}_{\pi\theta}[\nabla_{\theta}log\pi_{\theta}(s,a)\delta^{\pi\theta}]
\end{equation}
where $\mathbb{E}[\cdot]$ denotes the expected value, $s$ is the state in the state space $S$, $a$ is the action in the action space $A$ and $\delta$ is the temporal difference error.

\subsection{Training}
\begin{figure}[!t]
    \centering
    \includegraphics[width=3.4in]{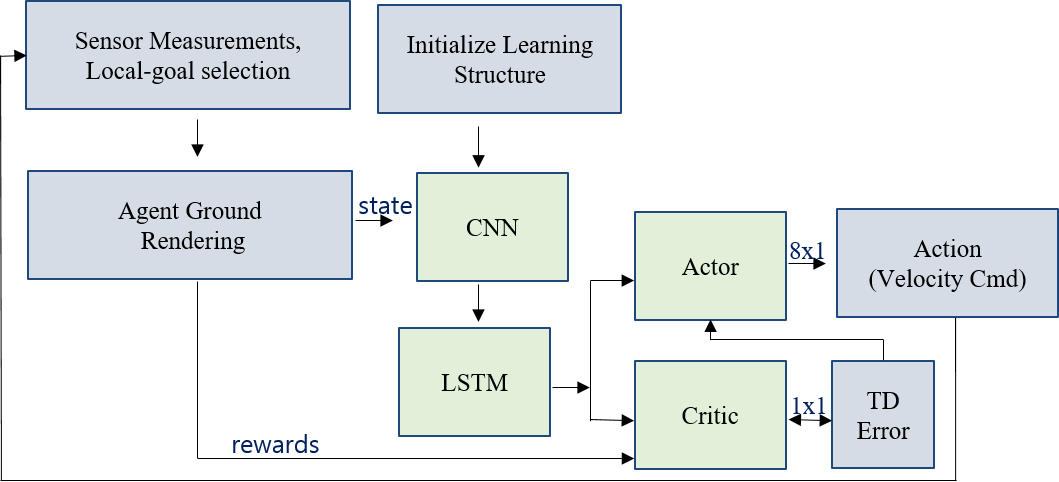}
    \caption{Overall training algorithm flow of proposed DRL based exploration scheme.}
    \label{fig:architecture}
\end{figure}

Training is performed in gazebo simulation environment. The sensor set and the size of a vehicle can be configured similar to the real world robot. As shown in Fig. \ref{fig:agent_ground}, the objective of this training step is to move the blue dot agent to the red dot local-goal. The black point is an obstacle, so if the agent reaches it, it will be penalized. Therefore, the agent must reach the red dot while avoiding obstacles to maximize the reward. Because the agent ground is configured with a resolution of 10cm, the local-goal from the agent is located 40 pixels ahead of the agent. The maximum number of steps per episode is set to 200 steps, and when the agent reached the target or obstacle, the episode is immediately terminated, rewarded, and initialized to a new episode.\\ Fig. \ref{fig:architecture} shows the overall training framework in block diagram representation which is following advantage actor-critic (A2C) algorithm Flow.

\subsection{Curriculum Learning}
\begin{figure}[!t]
    \centering
    \includegraphics[width=3.4in]{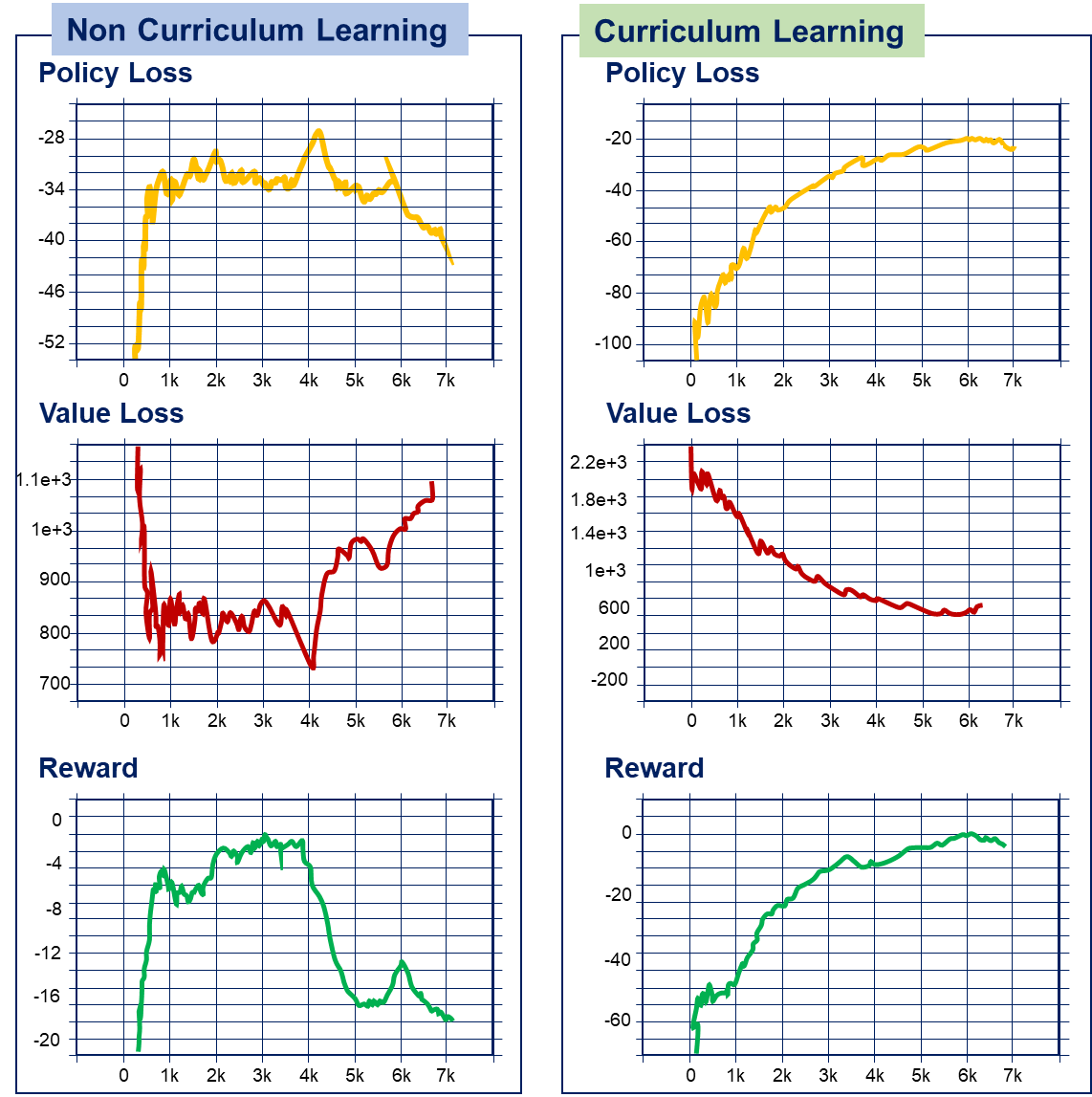}
    \caption{Training result of non-curriculum/curriculum learning. The introduction of curriculum learning confirms that the rewards and losses converge correctly}
    \label{fig:curriculum}
\end{figure}

If a local goal is placed far from the start at a time when no learning has been made, for example, aiming at a distance of 4 meters from the current position of the vehicle, the agent must take at least 40 consecutive successful actions to reach the local-goal. As a result of learning in this condition, the agent cannot reach the local-goal within the limited 200 steps. Accordingly, in order to somehow end the episode, the learning parameters were converged toward terminating the episode by hit the wall/obstacle (see Fig. \ref{fig:curriculum}-Non curriculum learning). Thus, to solve this problem, the method of curriculum learning \cite{bengio2009curriculum} is introduced. We set a goal of 2m distance up to the first 500 episodes, and learned 3m and 4m local goals for the rest of the training until 1500 episodes. The effects of curriculum learning were shown in Fig. \ref{fig:curriculum}-Curriculum learning. Policy loss, value loss and reward have been converging consistently as episodes increases.
\section{EVALUATION}
\subsection{Simulation}
\begin{figure}[!t]
    \centering
    \includegraphics[width=3.4in]{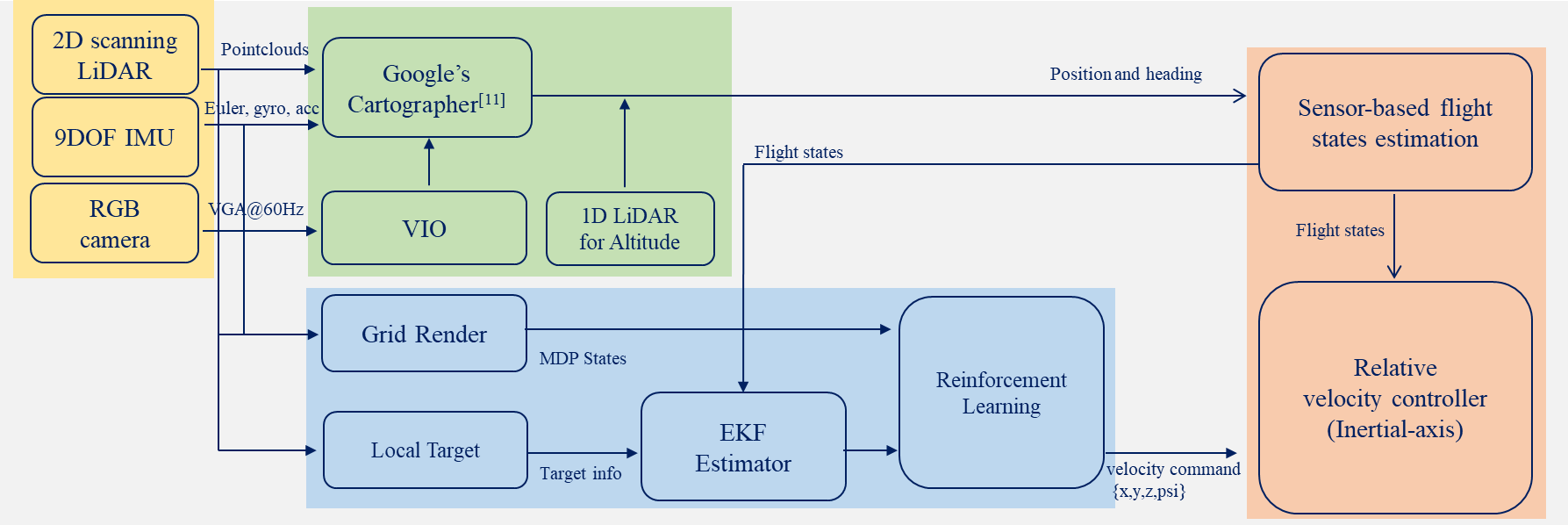}
    \caption{System overview of our DRL-based autonomous exploration aerial vehicle. Occupancy grid is obtained upon an existing open-source implementation \cite{hess2016real} for indoor localization purpose only}
    \label{fig:system}
\end{figure}

\begin{figure}[!t]
    \centering
    \includegraphics[width=3.4in]{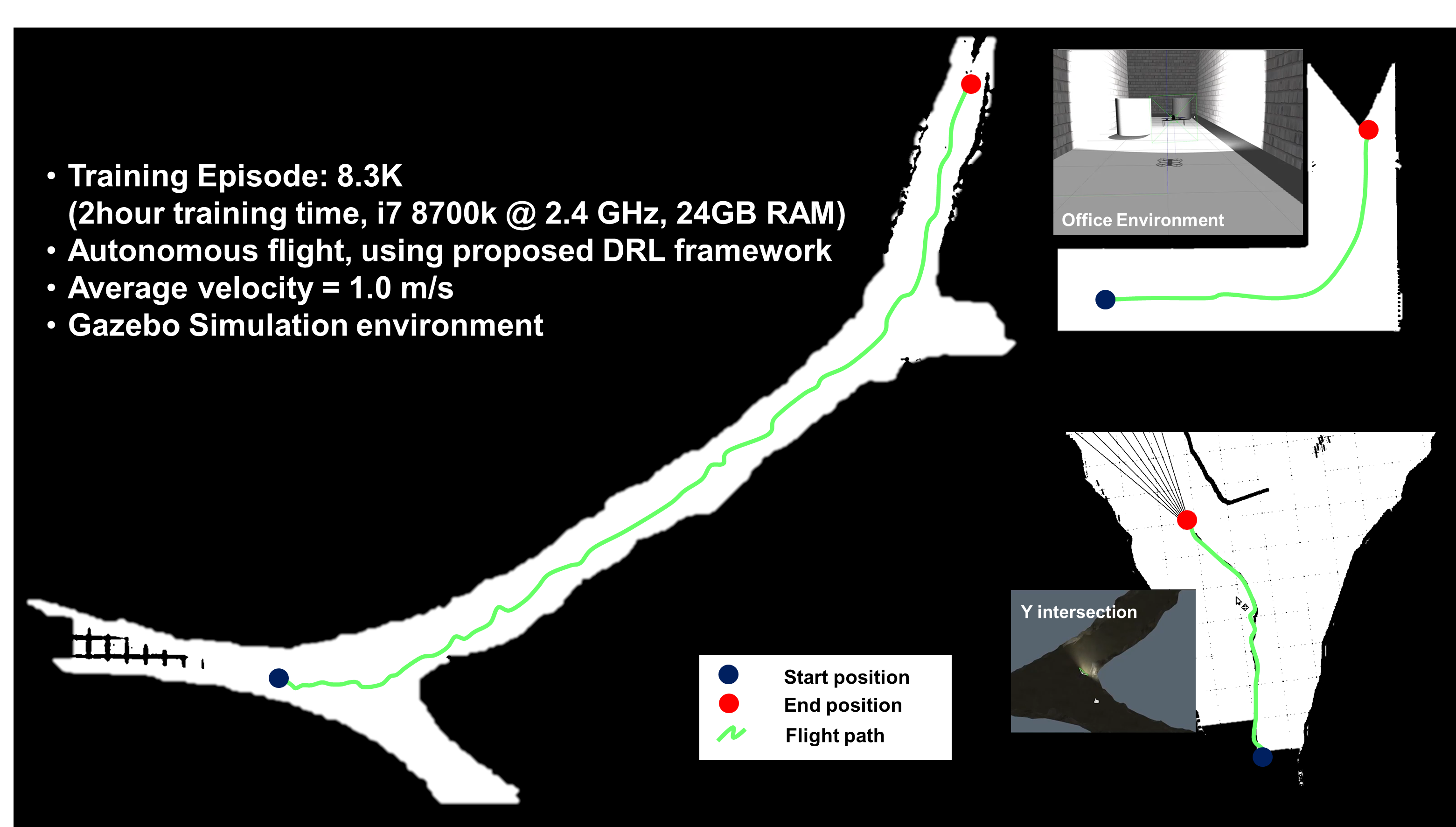}
    \caption{DRL-based autonomous exploration in simulation environment. The explored path shows well trained result by successfully avoiding collision with obstacles/walls}
    \label{fig:sim_eval}
\end{figure}

In order to evaluate the DRL-based autonomous motion planner, Various simulation environments is configured. For flight testing, the overall software architecture, including DRL techniques, is represented in Fig. \ref{fig:system}.\\
Fig. \ref{fig:sim_eval} shows DRL-based motion planner result in various simulation environments such as \textbf{"L"} shaped office environment, \textbf{"Y"} intersection and large-scale underground mines. In each environment, the proposed motion planner fly stably without collision. The exploration was expanded by the receding Horizon method while maintaining a constant interval from the current location of the Vehicle to the local-goal, resulting in successful flights of more than 200 meters in the underground mine simulation environment.
Additionally we compare computing costs of the traditional \textit{A*+ESDF} planner and the proposed \textit{DRL-based motion planner}. The traditional method is map-based navigation method, which produces grid-map from sensor information. After voxelizing it, the path is planned and the action is finally derived.
\begin{figure}[!t]
    \centering
    \includegraphics[width=3.4in]{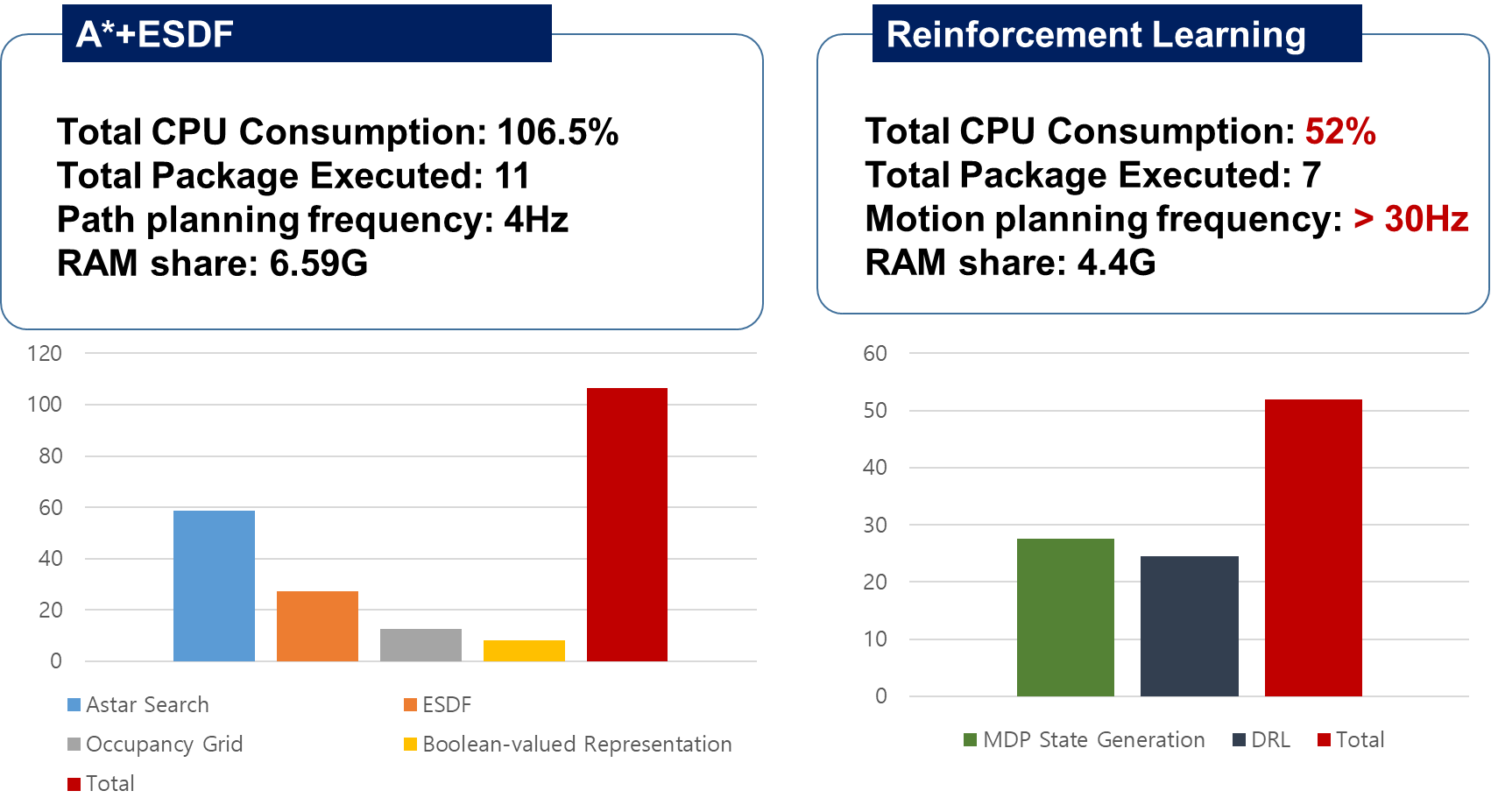}
    \caption{Comparison of computing cost between grid-based planning method and newly adopted DRL based motion planner. The proposed DRL-based planner consumes almost half of the energy of traditional path planner}
    \label{fig:computing_cost}
\end{figure}
\begin{figure*}[!t]
    \centering
    \includegraphics[width=7.0in]{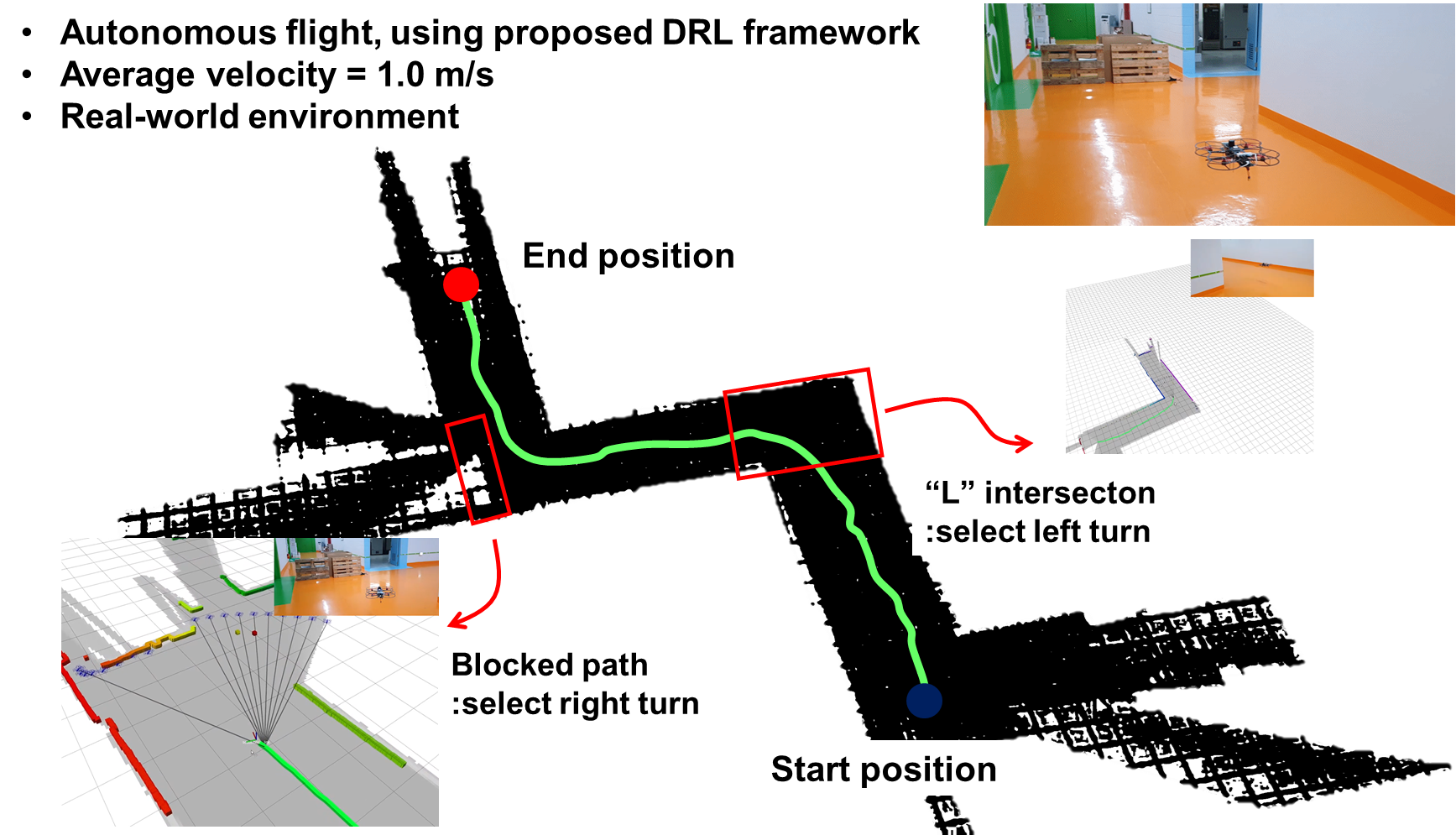}
    \caption{DRL-based motion planner, which flies about 70 meters in indoor underground environment. Flight paths obtained through actual flight experiments and flight scenes taken by external cameras are displayed together with capture images. Using the proposed DRL-based autonomous exploration method, the aerial robot fly on a safe path on a course consisting of various environments such as straight path, \textbf{"L"} intersection, and 3 way junction.}
    \label{fig:real-world}
\end{figure*}
On the other hand, the proposed DRL-based framework is a mapless-navigation method in which the action is derived immediately after the sensor information passes through the DRL-network. At this time, we decouple the action and control to reduce synthetic error caused by the dynamic effect of the system. Instead, we linked action with the Velocity command so that the results learned in simulation could be used directly for real-world regardless of low-level dynamics of the actual robot. According to Fig. \ref{fig:computing_cost} the CPU consumption rate decreased by about 50\% and the total number of packages running decreased by about 40\% compared to the traditional method. This indicates that the end-to-end method, the DRL technique, is more energy efficient than the traditional method.
\subsection{Real-world evaluation}
Our system is a 350-size frame with the DJI Snail racing drone propulsion system, which can lift all the sensor systems required for autonomous exploration purpose and fly as long as twelve minutes. We chose the Px4 autopilot for low-level flight control, and NVIDIA Jetson TX2 GPU for high-level mission management. The TX2 module is mounted on a third-party carrier board (Auvidea J120). For altitude measurement, the drone is equipped with a Terabee TeraRanger-Evo one-dimensional LiDAR. The 2D SLAM and the altitude information are sent to the Extended Kalman Filter (EKF) running on the NVIDIA Jetson TX2 to compute more accurate state estimation by fusing with Microstrain 3DM-GX4-25 Inertial Measurement Unit (IMU).\\
Fig. \ref{fig:real-world} shows the navigation of unexplored corridors using the proposed MDP state generalization and local-goal selection methods. In the corridor environment consisting of a straight section and a corner section of varying width, it can be confirmed that the flight is successfully performed by the proposed method. The path is also created in the middle of the corridor, showing the robot moved properly to the collision-free path.
\section{CONCLUSIONS}
\label{C_FW}
In this paper, we try to solve the problem of path planning by the deep reinforcement learning method. First, we express the state of the Markov decision process, which excludes dynamics of the hardware system (vehicle, sensor, etc.) as much as possible. Through this, we want to use the policy learned in simulation directly without additional training in real-world. The rewards are designed to receive -100 points in the event of a collision with an obstacle/wall and +10 points if the agent reached the local-goal. If the agent is moving around the ground, the difference between the Euclidean distance from the previous position to the goal and the current position to the goal was designed as a reward. It is designed to receive a reward of + when the agent moves closer to the goal and - when the agent moves away.\\
We first tried to learn in this way, but the values of rewards and total steps did not converge easily. Thus, the method of curriculum learning was introduced. The curriculum learning method is also called self-paced learning as a way to teach from easy methods step by step so that more difficult problems can be solved. The position of the target point that the agent had to reach increased from 2m to 5m as the episode increased, resulting in a successful learning result. These learning results were first validated in gazebo simulations. The reason for choosing gazebo as the simulation environment was useful for this high-level control study because it could use the same controller of the px4 stack used as the low-level controller of the aerial-robot. Also, sensor data are highly correlated so that simulation results can be easily applied as real-world. Through Gazbo simulation, we conducted learning in office environment, straight course environment, and subterranean environment. As a result of deploying the learned policy, it is possible to successfully explore the unknown environment. These are applied directly to the real-world without further learning. The flight test is conducted in the basement of KAIST's KI building and is very successful in exploratory flights.\\
In this paper, we use 2D LiDAR point cloud. In other words, we do the motion plan using only point cloud that is projected to a horizontal plane. For Future work, we plan to use three-dimensional LiDAR to project point cloud into a vertical plane and create another DRL module to plan the motion in vertical plane also to focus on solving the Full-3d problem.
\bibliographystyle{IEEEtran} 
\bibliography{references}
\end{document}